# Enhancing Genetic Algorithms using Multi Mutations: Experimental Results on the Travelling Salesman Problem

Ahmad B. A. Hassanat [1*], Esra'a Alkafaween [2], Nedal A. Al-Nawaiseh [3], Mohammad A. Abbadi [4], Mouhammd Alkasassbeh [5], and Mahmoud B. Alhasanat [6]

[1,2,4,5] IT Department, Mutah University, Mutah, Karak, Jordan.
[3] Department of Public Health and Community Medicine, Mutah University, Mutah, Karak, Jordan.
[6] Department of Civil Engineering, Al-Hussein Bin Talal University, Maan, Maan, Jordan.
[*] Corresponding Author: Ahmad B. A. Hassanat, Mutah Street, Mutah, Karak, 61711, Jordan, Email address: Ahmad.hassanat@gmail.com

*Abstract*—Mutation is one of the most important stages of genetic algorithms because of its impact on the exploration of the search space, and in overcoming premature convergence. Since there are many types of mutations one common problem lies in selecting the appropriate type. The decision then becomes more difficult and needs more trial and error to find the best mutation to be used.

This paper investigates the use of more than one mutation operator to enhance the performance of genetic algorithms. New mutation operators are proposed, in addition to two election strategies for the mutation operators. One is based on selecting the best mutation operator and the other randomly selects any operator.

Several experiments were conducted on the Travelling Salesman Problem (TSP) to evaluate the proposed methods. These were compared to the well-known exchange mutation and rearrangement mutation. The results show the importance of some of the proposed methods, in addition to the significant enhancement of the genetic algorithms' performance, particularly when using more than one mutation operator.

*Index Terms*— Mutation operator, Nearest Neighbor, Multi Mutations, TSP, GA, AI, Evolutionary Computation

## I. INTRODUCTION

GENETIC algorithms (GA) are adaptive heuristic random search techniques [1], and are a sub-family of evolutionary algorithms that mimic the theory of evolution and natural selection. The basic principles of genetic algorithm were presented by John Holland in the 1970s [2]. The effectiveness of genetic algorithms has been proven by solving many optimization problems [3], [4] and [5].

There are many applications of genetic algorithms in various areas, such as image processing [6], software engineering [7], computer networks [8], robotics [9], and speech recognition [10].

Genetic algorithms are concerned, in general, with how to produce new chromosomes (individuals) that possess certain features through recombination (crossover) and mutation operators. Therefore, individuals with appropriate characteristics have the strongest chance of survival, while individuals with inappropriate characteristics are less likely to survive. This simulates Darwin's theory of evolution by natural selection, colloquially described as survival of the fittest [11], [12] and [13].

Typically, GAs start with a number of random solutions (initial population). These solutions are encoded according to the current problem, forming a chromosome for each individual (solution). The quality of each individual is then evaluated using a fitness function, after which the current population changes to a new population by applying three basic operators: selection, crossover and mutation. The efficiency of a genetic algorithm is based on the appropriate choice of these operators and strategy parameters [14] associated with ratios, such as crossover ratio and mutation ratio [15]. Many researchers have shown the effect of the two operators—crossover and mutation—on the success of the GA, and where success lies in both, whether crossover is used alone or mutation alone or both, as in [16] and [17].

One of the common issues with genetic algorithms is premature convergence [18] which is directly related to the loss of diversity [19]. Achieving population diversity is a desired goal, as the search space becomes better (diverse) accordingly, and also avoids a suboptimal solution. According to Holland, mutation is considered an important mechanism to maintain diversity [20]. Researchers [21], explored new areas in the search space, thus avoiding the convergence of the local optimum [22]. The need for mutation is to prevent loss of genetic material where the crossover does not guarantee access to new parts of the search space [23]. Therefore, random changes in the gene through mutation helps provide diversity in the population [15].

Genetic algorithms have evolved from what was prevalent in the era of Holland [24]. Classical mutation (bit-flip mutation) developed by Holland with different encoding problems no longer fits for TSP because it is difficult to encode a TSP as a binary string that does not have ordering dependencies [25]. Therefore, several types of mutation of various types of encoding have been proposed, including Exchange Mutation [26], Displacement Mutation [27], Uniform Mutation and Creep Mutation [28], Inversion Mutation [29], etc. One problem lies in our selection of which type(s) to use to solve a specific problem, which increases the difficulty in our decision and requiring more trial and error to find the best mutation to be used. To overcome this problem, several researchers have developed new types of GA that use more than one mutation operator at the same time [30], [31] and [32]. This paper contributes to previous work to overcome the problem of determining which mutation to use.

The contribution of this paper is two-fold: (1) proposals of new mutation operators for TSP, and (2) investigations into the effect of using more than one of these mutations on the performance of the GA.

The rest of this paper presents some of the related previous work and the proposed methods. This paper also discusses the experimental results, which were designed to evaluate the proposed methods. Conclusions and future work are presented at the end of the paper.

## II. RELATED WORK

To increase the effectiveness of the algorithm in tackling a problem, researchers have focused on improving the performance of Genetic Algorithms to overcome premature convergence.

Soni and Kumar studied many types of mutations that provide approximate solutions to the TSP [28]. Larrañaga et al. presented a review of how to represent travelling salesman problems and the advantages and disadvantages of different crossover and mutation operators [25]. Louis and Tang proposed a new mutation called greedy-swap mutation, so that two cities are chosen randomly in the same chromosome, and switching between them if the length of the new tour obtained is shorter than the previous ones [33].

Hong et al. proposed an algorithm called the Dynamic Genetic Algorithm (DGA) to simultaneously apply more than one crossover and mutation operator. This algorithm automatically selects the appropriate crossover and appropriate mutation, and automatically adjusts the crossover and mutation ratios, based on the evaluation results of the respective offspring in the next generation. In comparing this algorithm with the simple genetic algorithm that commonly uses one crossover process and one process of mutation, the results showed the success of the proposed algorithm in performance [30].

Deep and Mebrahtu proposed an Inverted Exchange mutation and Inverted Displacement mutation, which combine inverted mutation with exchange mutation and combines inverted mutation with displacement mutation. The experiment was performed on the TSP problem and the results were compared with several existing operators [23].

Hong et al. proposed a Dynamic Mutation Genetic Algorithm (DMGA) to simultaneously apply more than one mutation to generate the next generation. The mutation ratio is also dynamically adjusted according to the progress value that depends on the fitness of the individual. This decreases the ratio of mutation if the mutation operator is inappropriate, and vice versa, increases the ratio of mutation if the operator is appropriate [34] [31]. Dynamically adjusting the mutation ratio was studied and used later by several researchers [ [35] and [36]].

Hilding and Ward proposed an Automated Operator Selection (AOS) technique which eliminated the difficulties that appear when choosing crossover or mutation operators for any problem. In this technique, they allowed the genetic algorithm to use more than one crossover and mutation operators; taking advantage of the most effective operators to solve problems. The operators were automatically chosen based on their performance, and thereby reducing the time spent choosing the most suitable operator. The experiments were performed on the 01-knapsack problem. This approach was more effective as compared to the traditional genetic algorithm [32].

Dong and Wu proposed a dynamic mutation probability, which calculates the mutation rate by the ratio between the fitness of the individual and the most fit in the population. This ratio helps the algorithm to avoid local optima and also leads to the population's diversification [37]. Patil and Bhende presented a study of the various mutation-based operators in terms of performance, improvement and quality of solution. A comparison was made between Dynamic Mutation Algorithm, Schema Mutation Genetic Algorithm, Compound Mutation Algorithm, Clustered-based Adaptive Mutation Algorithm, and Hyper Mutation-Based Dynamic Algorithm [38].

## III. METHODS

Many researchers have attempted to prevent local convergence in different ways. Since mutation is a key operation in the search process, we found several mutation methods in the literature. The question is: what is the best method to use? To answer this question, and in the hope of avoiding local optima and increasing the diversification of the population, we have proposed and implemented 10 types of mutations to be compared with two of the well-known types, namely, Exchange mutation and Rearrangement mutation [39].

In the following we describe each operator. It is important to note that mutation methods described next subsections were designed specifically for the TSP problem. However, they can be customized to fit other problems, such as the knapsack problem with special treatment that goes with the definition of the problem and the designed chromosome.

*A. Worst gene with random gene mutation (WGWRGM)*

To perform this mutation, we need to search for the "worst" gene in the chromosome from index 0 to L-1, where L is the length of the chromosome. The worst gene varies depending on the definition of the worst for each problem and each method. Basically, the worst gene is the point in a specific chromosome that contributes the maximum to increase the cost of that chromosome (solution).

In this method, the worst gene in the TSP's chromosome is the city with the maximum distance from its left neighbor, while the worst gene in the knapsack problem is the point with the lowest value-to-weight ratio, and so on. The worst gene is defined based on the definition of the problem.

After identifying the worst gene for a TSP chromosome, another gene is randomly selected, and then both genes are swapped, as in the Exchange mutation. In the knapsack problem, however, the worst gene is not swapped with a random gene but removed from the solution (converted to zero in the binary string), and another random (zero) gene is converted to one, to hopefully create a better offspring. Figure 1 shows an example of WGWRGM.

The worst gene (WG) can be calculated for a minimization problem such as TSP using:
$$WG = \underset{1 \leq i < n}{\mathrm{argmax}}(Distance(C[i], C[i+1])) \quad (1)$$
and for the maximization problem, such as the knapsack problem using:
$$WG = \underset{0 \leq i < n}{\mathrm{argmin}}\left(\frac{Value(C[i])}{weight(C[i])}\right) \quad (2)$$
where C represents the chromosome, *i* is the index of a gene within a chromosome, and the distance function for the TSP can be calculated using either Euclidian distance or the distances table between cities. In the case of TSP, searching for the WG starts at index 1, assuming that the route-starting city is located at index 0, while this is not the case for other problems such as the knapsack problem (Equation 2).

The previous equations are used for the chromosome, and the worst gene of this chromosome that exhibits the maximum distance is used for the mutation operation.

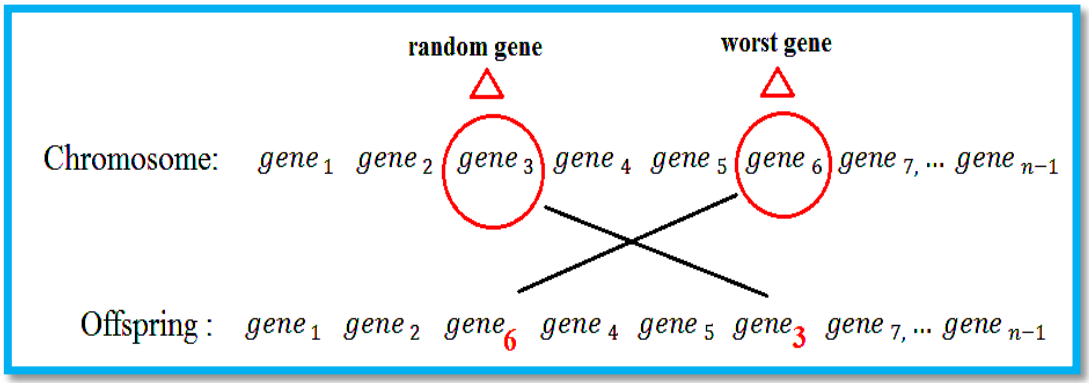

Fig. 1. Example of WGWRGM

**Example 1**. Example of applying WGWRGM to a specific chromosome of a particular TSP, the measurements are in centimeters (cm) to demonstrate the example on a printed paper. Suppose that the chromosome chosen for mutation is:

CHR1: A→B→E→D→C→A, as depicted in Figure 2(a).

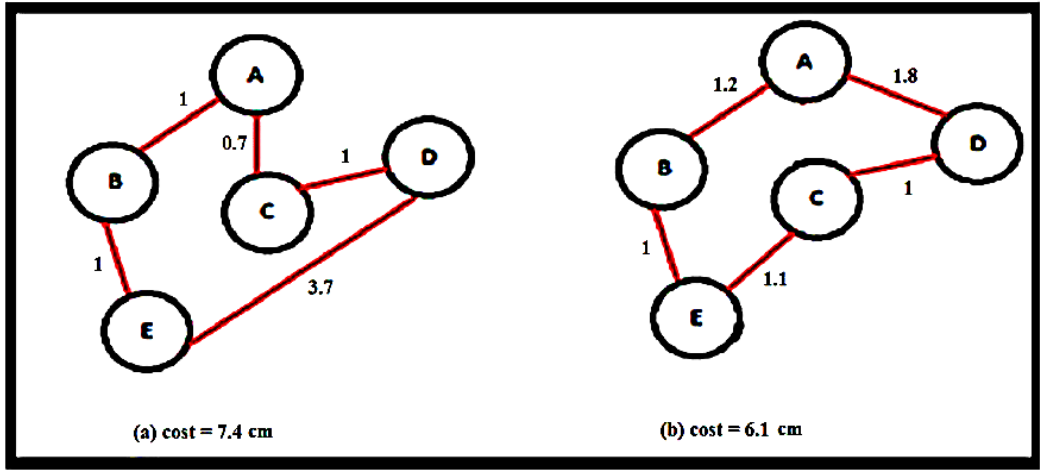

Fig. 1. Example of applying WGWRGM to a specific chromosome of a particular TSP

To apply WGWRGM:
    Step 1: Find the worst gene in the parent. According to Figure 2, the worst gene is (D).
    Step 2: Suppose that the city which has been selected at random is (C).
    Step 3: Apply the Exchange mutation in this chromosome by swapping the positions of the two cities (see Figure 2(b)).
    The output offspring becomes: A→B→E→C→D→A.

*B. Worst gene with worst gene mutation (WGWWGM)*

Although this type is similar to the WGWRGM, the difference is searching for the two worst genes then exchange positions of both the selected genes with each other. Finding both worst genes is similar to finding the two maximum values in a single array, if the problem being dealt with is a minimization problem. For the maximization problem, the algorithm of finding the two minimum values is used. The definition of the worst gene concept is different from one problem to another. For example, the two worst genes in the knapsack problem can be found by applying Equation (2) twice. Instead of swapping, both become zeros and two random (zeros) genes become ones. Figure (3) shows a TSP example of the WGWWGM.

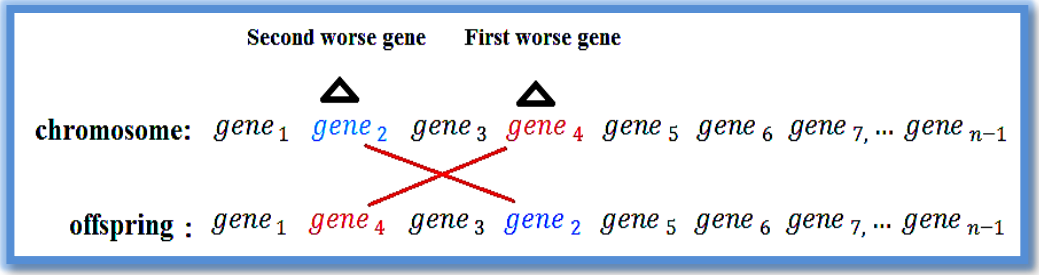

Fig.2. Example of WGWWGM

*C. Worst left and right gene with random gene mutation (WLRGWRGM)*

This method is also similar to the WGWRGM but the difference is that the worst gene is the one with the maximum total distance between that gene and both of its neighbors—the left and the right neighbors. Considering both distances (left and right) might be more informative than considering only one distance from left or right.

The worst gene (WLRgene) can be calculated for the TSP using:

$$W_{LRgene} = \underset{1 \leq i < n-2}{\mathrm{argmax}}(Distance(C[i], C[i-1]) + Distance(C[i], C[i+1])) \qquad (3)$$

and if it is a maximization problem using:

$$W_{LRgene} = \underset{1 \leq i < n-2}{\mathrm{argmin}}(Distance(C[i], C[i-1]) + Distance(C[i], C[i+1])) \qquad (4)$$

Equation (3) can be used for minimization problems, and Equation (4) for maximization problems, e.g. finding the maximum route in TSP. The extreme genes, the first and last ones in a chromosome, can be handled in a circular way, i.e. the left of the first gene is the last gene.

The worst gene for minimization problems is the one that the sum of the distances with its left and right neighbors is the maximum among all genes within a chromosome; and vice versa for Maximization problems. In this mutation, the position of the worst gene is altered with the position of another gene chosen randomly.

This mutation is not defined for the knapsack problem, as the distance is not defined for such a problem.

**Example 2**. Example of applying WLRGWRGM to a specific chromosome of a particular TSP, Figure 4(a) represents the chromosome chosen for mutation, which is: A→B→E→H→F→D→C→A.

According to Figure 4 (a), the WLRgene is city D because the total distance from city D to city F and from city D to city C is the maximum distance (4.5 cm). If randomly choosing city H to swap with the WLRgene, the output offspring after applying WLRGWRGM mutation is A→B→E→D→F→H→C→A (see Figure 4(b)).

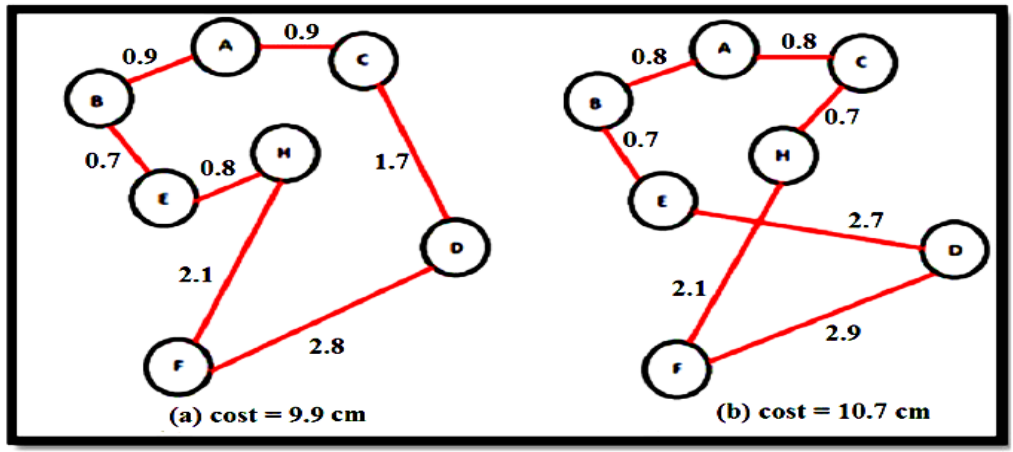

Fig.3. Example of applying WLRGWRGM on a specific chromosome of particular TSP

As can be seen from Figure 4, the new offspring does not provide a better solution which is true for many mutations. Due to randomness, there is no guarantee for better offspring all the time.

*D. Worst gene with nearest neighbor mutation (WGWNNM)*

This method uses the idea of the nearest neighbor city, which provides an heuristic search process for mutation. Basically, the worst gene is swapped with one of the neighbors of its nearest city. The WGWNNM is performed as follows:

> *Step* 1: Search for the gene (city) in a tour characterized by the worst with its left and right neighbors (WLRgene) as in WLRGWRGM mutation. This city is called the worst city.
> *Step* 2: Find the nearest city to the worst city (from the graph) and call it Ncity. Then search for the index of that city in the chromosome and call it Ni.
> We need to replace the worst city with another one around the Ncity other than the Ncity itself. The term around is defined by a predefined range, centered at the Ncity. To give the algorithm some degree of randomness, the algorithm arbitrarily used (Ni ± 5) as a range around the index of the Ncity. The out-of-range problem with the extreme points is solved by dealing with the chromosome as a circular structure.
> *Step* 3: Select a random index within the range. The city at that index is called random city.
> *Step* 4: Swap between the worst city and the random city.

Example 3. Example of applying WGWNNM to a specific chromosome of a particular TSP, suppose that the chromosome chosen for mutation is: A→B→F→D→E→C→H→A, as depicted in Figure 5(a).

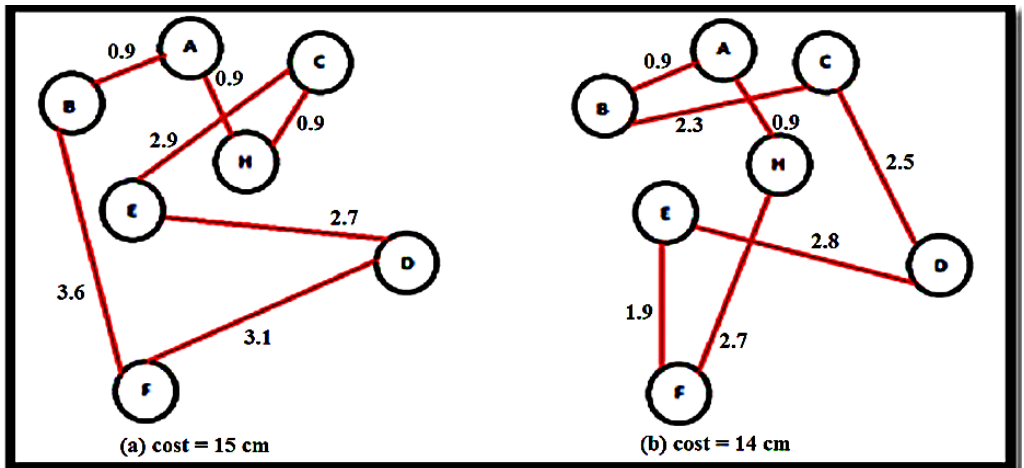

Fig. 4. Example of applying WGWNNM to a specific chromosome of particular TSP

By applying WGWNNM:
    *Step* 1: Find the WLRgene in the chromosome. According to the graph, the worst city is F (6.7 cm).
    *Step* 2: Find the nearest city to the worst city, which is E according to the distance table. This city is called Ncity.
    *Step* 3: Search for a city around Ncity at random in the range ± 5. Suppose we choose city C.
    *Step* 4: Apply the Exchange mutation in this chromosome by swapping the position of the two cities F and C (see Figure 5(b)). The output offspring is A→B→C→D→E→F→H→A.

This mutation cannot be defined for the knapsack problem, as the nearest neighbor approach is not defined for such a problem.

*E. Worst gene with the worst around the nearest neighbor mutation (WGWWNNM)*

This mutation is similar to the WGWNNM but the only difference is in the selection of the swapped city. The swapped city is not randomly selected around the nearest city as in WGWNNM, but rather is chosen based on its distance from the nearest city. By considering the furthest city from the nearest city to be swapped with the worst city, this brings nearest cities together, and sends furthest cities far away.

This mutation will hopefully provide better offspring. However, there is no guarantee, as the swapped furthest city might be allocated in a place neighboring very far away cities, which creates a new offspring with longer TSP route.

The WGWWNNM is also cannot be defined for the knapsack problem, as the distance is not defined for such a problem neither the nearest neighbor approach.

*F. Worst gene inserted beside nearest neighbor mutation (WGIBNNM)*

This type of mutation is similar to the WGWNNM, after finding the indices of the worst city and its nearest city. The worst city is moved to be a neighbor to its nearest city, and the rest of the cities are then shifted either left or right depending on the locations of the worst city and its nearest city.

In other words, if the worst city was found to the right of its nearest city, the worst city is moved to the left of its nearest city, and the other cities are shifted to the right of the location of the worst city. If the worst city was found to the left of its nearest neighbor, the worst city is moved to the location prior to the location of its nearest city, and the rest of the cities between this location and the previous location of the worst city are shifted to the right of that location, and vice versa.

**Example** 4. Example of applying WGIBNNM to a specific chromosome of a particular TSP, suppose that the chromosome chosen for mutation is: A→B→F→D→E→C→H→A, as depicted in Figure 5(a). By applying WGIBNNM:
    *Step* 1: Find the WLRgene in the chromosome. According to the graph, the worst city is F (6.7 cm).
    *Step* 2: Find the nearest city to the worst city, which is E according to the distance table. This city is called Ncity.
    *Step* 3: Now F is moved prior to E, and (A and B) are shifted right to get a new chromosome A→B→D→ F→E→C→H→A.

*G. Random gene inserted beside nearest neighbor mutation (RGIBNNM)*

This mutation is almost the same as the WGIBNNM, except that the worst city is selected randomly, i.e. the worst city concept here is not defined, it is just a random city, and is not based on its negative contribution to the fitness of the chromosome. We propose the RGIBNNM to enhance the performance of the WGIBNNM by enforcing some randomness to increase diversity in the search space.

The RGIBNNM is also cannot be defined for the knapsack problem, as the distance is not defined for such a problem neither the nearest neighbor approach.

*H. Swap worst gene locally mutation (SWGLM)*

This mutation is based on finding the worst gene using WLRGWRGM, then it swaps related genes locally, either the left neighbors are swapped, or the worst gene is swapped with its right neighbor. The best resulting offspring decides which genes will be swapped. This mutation is summarized as follows:

*Step* 1: Search for the worst gene, the same as for WLRGWRGM.
*Step* 2: Swap the left neighbor of the worst gene with its left neighbor, and calculate the fitness (C1) of the new offspring (F1).
*Step* 3: Swap the worst gene with its right neighbor, and calculate the fitness (C2) of the new offspring (F2).
*Step* 4: If C1 > C2, then return F2 as the legitimate offspring and delete F1, otherwise return F1 as the legitimate offspring and delete F2 (see Figure 6).

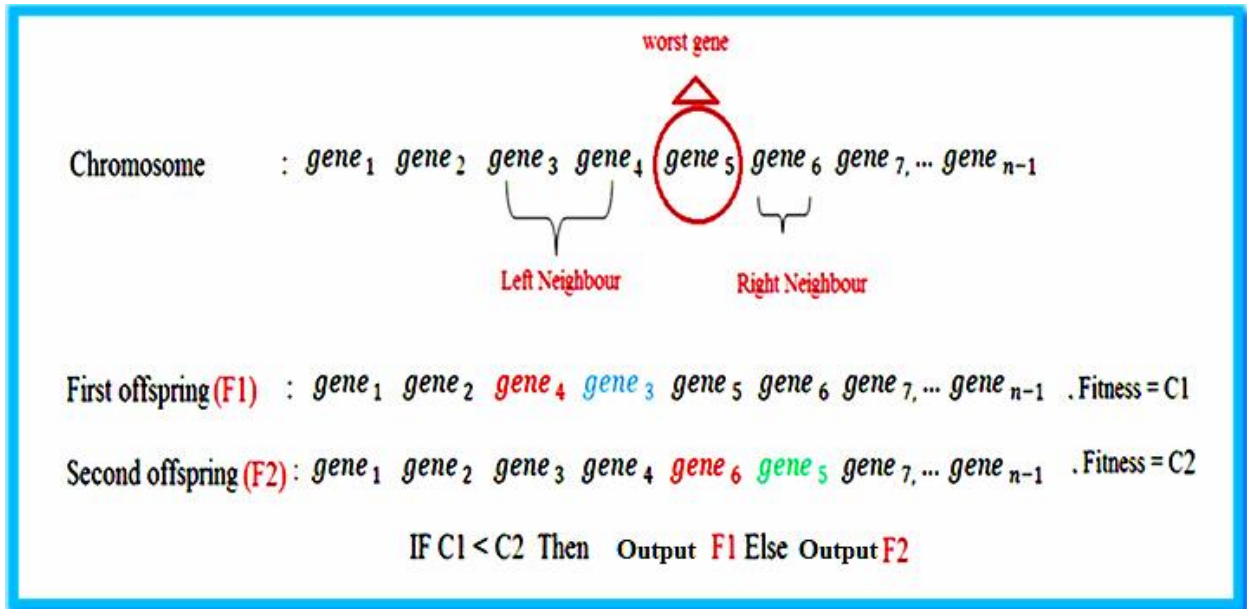

Fig. 5. Example of SWGLM

Example 5. Example of applying SWGLM to a specific chromosome of a particular TSP, suppose that the chromosome chosen for mutation is: A→B→F→E→H→D→C→A, as depicted in Figure 7(a). To apply SWGLM:

*Step* 1: Find the worst gene in the chromosome. According to the graph, the worst city is E (6.2 cm).
*Step* 2: Swap the two left neighbors of E, which are B and F. The first offspring become A→F→B→E→H→D→C→A, and the cost of this offspring is C1 (15 cm) (see Figure 7(b)).
*Step* 3: Swap between worst city E and its right neighbor H. The second offspring become A→B→F→H→E→D→C→A. The cost of this offspring is C2 (10.2 cm) (see Figure7(c)).
*Step* 4: Compare the cost (C1, C2) and the least among them is the output offspring.

Based on the graph the output offspring is A→B→F→H→E→D→C→A (Figure 7(b)).

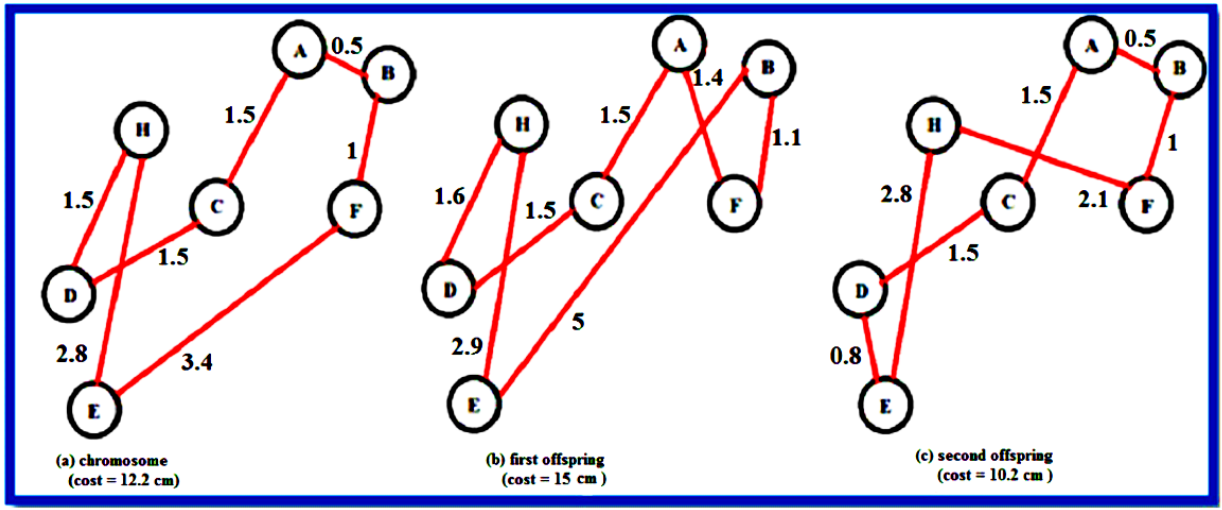

Fig. 6. Example of applying SWGLM to a specific chromosome of particular TSP

*I. Insert best random gene before worst gene mutation (IBRGBWGM)*

This method is based on finding the worst gene, as in WGWRGM, which is the city with the maximum distance from its left neighbor. Choose a random number of cities, insert the one with the minimum distance to both the worst city and its left neighbor between them. This mutation is summarized as follows:

*Step* 1: Search for the city that is characterized by the worst city as in WGWRGM and find the index of its previous city.
*Step* 2: Select a certain number of random cities. In this work we chose five random cities arbitrarily excluding the worst city and its previous neighbor (PN).
*Step* 3: For each random city calculate the distance to the worst city (D1) and the distance to PN (D2).
*Step* 4: Find the best city from the random cites, which is the one with the minimum distance (D1+D2).
*Step* 5: Move the best city and insert it between the worst city and PN.
*Step* 6: Shift cities which are located between the old and the new location of the best city to legitimize the chromosome.

**Example** 6. Example of applying IBRGBWGM to a specific chromosome of a particular TSP, Figure 8(a) represents the chromosome chosen for mutation, which is: Chromosome: A→B→E→D→C→A.

According to Figure 8(a), the worst gene is city E because the distance to its left equals four centimeters. According to the graph, the best city is C—distance (C, E) + distance (C, B) is the minimum. The output offspring after applying the IBRGBWGM mutation is A→B→C→E→D→A (see Figure 8(b)).

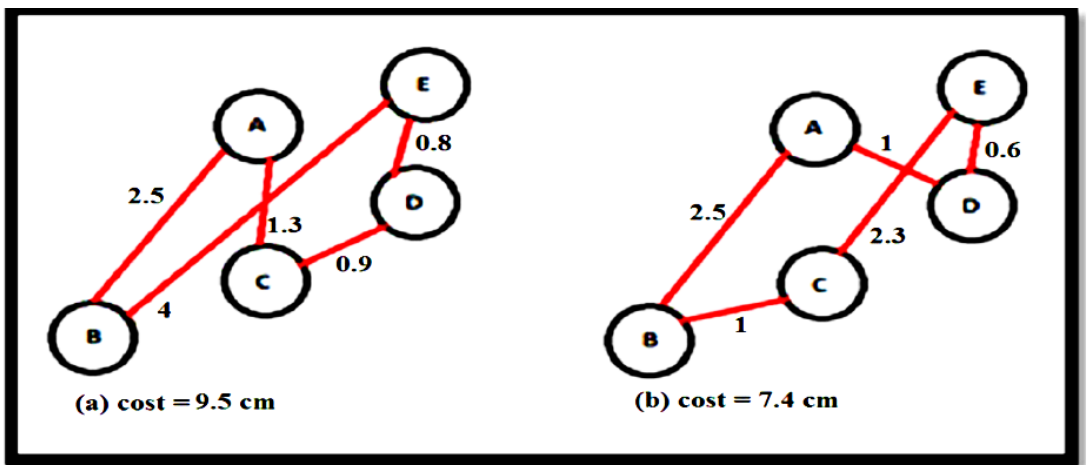

Fig. 7. Example of applying IBRGBWGM to a specific chromosome of particular TSP

*J. Insert best random gene before random gene mutation (IBRGBRGM)*

Sometimes the worst gene is located in the best possible location, thus swapping it with another gene might yield weak offspring. Therefore, it is important to have another mutation which does not depend on finding the worst gene but instead uses a random gene. This mutation is similar to IBRGBWGM, however, the difference is that the worst city is not chosen based on any distance but is instead chosen randomly to impose some diversity among the new offspring.

Another important motivation for proposing IBRGBRGM is the computation time. As with finding the worst gene, enforce a linear computation time along the chromosome—O(n) where n is the length of the chromosome. Finding the nearest neighbor approach also exhibits O(n) time complexity, while choosing a random gene takes only O(1). Finding the nearest gene from a constant (k) number of randomly selected genes takes O(k), which is approximate to O(1) when n (number of the cities in a TSP instance) is very large.

### K. Multi Mutation Operators Algorithms

A traditional genetic algorithm normally uses just one mutation operator. We propose using more than one mutation operator. Those different mutations are supposed to lead to different directions in the search space, thus increasing diversity in the population, and therefore improving the performance of the genetic algorithm. To do this we opted for two selection approaches: the best mutation, and a randomly chosen mutation.

#### 1) Select the best mutation algorithm (SBM)

This algorithm simultaneously applies multiple mutation operators to the same chromosome. To prevent duplication, it only considers the best offspring that is not found in the population to add to the population.

In this work, we defined 10 mutations to apply. The SBM implements the entire aforementioned methods—WGWRGM, WGWWGM, WLRGWRGM, WGWNNM, WGWWNNM, WGIBNNM, RGIBNNM, SWGLM, IBRGBWGM and IBRGBRGM—one after the other with each mutation producing one offspring. The best offspring that does not already exist in the population is added. In TSP the best offspring is the one with the minimum TSP route.

Using such a diverse collection of mutations anticipates that such processes encourage diversity in the population, thus avoids convergence to local optima and provides better final solutions.

#### 2) Select any mutation algorithm (SAM)

This algorithm tries to apply a mutation each time, which is selected from a collection of operators. The selection strategy is random. Each operator has the same probability to be chosen. The algorithm randomly chooses one of the aforementioned mutations each time it is called by the GA. Therefore, in each generation different mutations are chosen. This means that there is a different direction of the search space which is what we are aiming for; increasing diversity and attempting to enhance the performance of the genetic algorithm.

## IV. EXPERIMENTS AND DISCUSSIONS

To evaluate the proposed methods, we conducted two sets of experiments on different TSP problems. The aim of the first set of experiments was to examine convergence to a minimum value of each method separately. The second set of experiments was designed to examine the efficiency of the SBM and SAM algorithms and compare their performance with the proposed mutation operators—WGWRGM, WGWWGM, WLRGWRGM, WGWNNM, WGWWNNM, WGIBNNM, RGIBNNM, SWGLM, IBRGBWGM and IBRGBRGM—using the TSPLIB, a collection of travelling salesman problem datasets maintained by Gerhard Reinelt at http://comopt.ifi.uni-heidelberg.de/software/TSPLIB95/. The results of these experiments were compared with two existing mutations: Exchange mutation [26], and Rearrangement mutation [39].

In the first set of experiments, the mutation operators were tested using three test instances taken from TSPLIB [40], consisting of berlin52, ch130 and a280, each consisting of 52, 130, and 280 cities respectively.

The genetic algorithm parameters used are as follows:
1) Population size = 100.
2) The probability of crossover = 0.0.
3) Mutation's probability = 1.0.
4) The selection strategy is based on keeping the best k solutions, whether they are old parents or offspring resulted from the mutation operator(s), where k is the constant size of the population.
5) The termination criterion is based on a fixed number of generations reached. In our experiments the maximum number of generations = 1,600.
6) The chromosome used is a string of random sequence of cities' numbers, thus, the chromosome length is associated with the problem size n, which is the number of cities for each TSP problem.

The GA was applied 10 times using each of the proposed mutation, the average of the best solutions from the 10 runs, for each generation, for each method, for each TSP instance was recorded, starting from generation 1 up to generation 1,600.

Results from the first test indicate that the best performance was recorded by the SBM, followed by the SAM. This compared well with the rest of the mutations because it showed good convergence to a minimum value.

The efficiency of each of the 14 mutations (10 proposed, 2 from the literature, and 2 selection strategies) is shown in Figures 9-11. A closer look at these figures reveals that the SBM and SAM algorithms outperform all other methods in the speed of convergence.

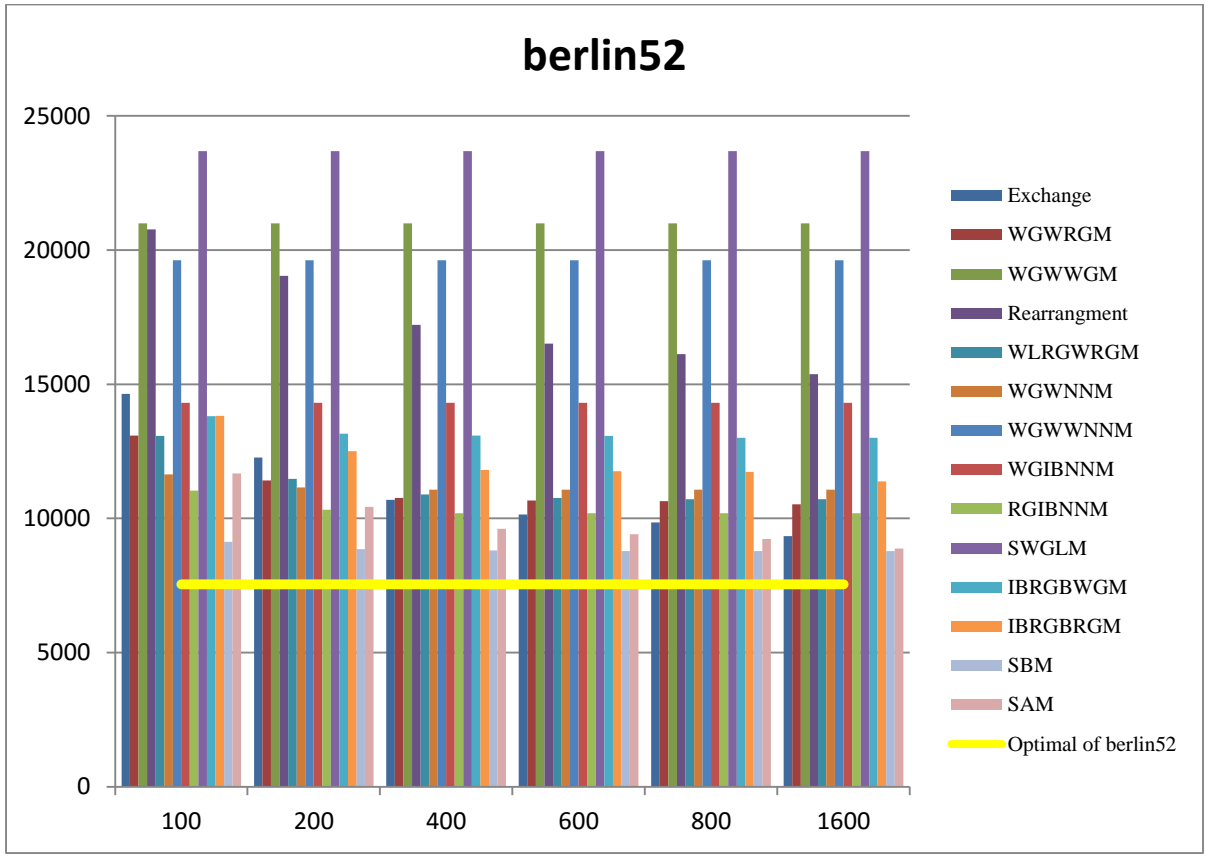

Fig. 8. Mutation's convergence to the minimum value, TSP (berlin52)

As seen in Figure 9, the results indicate the efficiency of the SBM and SAM algorithms, where the speed of convergence of a near optimal solution with the progress of the generations is faster than the use of a certain type of mutation alone. The Exchange mutation followed by RGIBNNM also showed the extent of their influence on the quality of the solution.

One result in Figure 10 indicates that the SBM algorithm showed faster convergence to the minimum value followed by SAM, and these algorithms showed better performance than the remaining mutations. At the level of mutation alone, the WLRGWRGM mutation followed by WGWRGM showed a better performance than the other mutations.

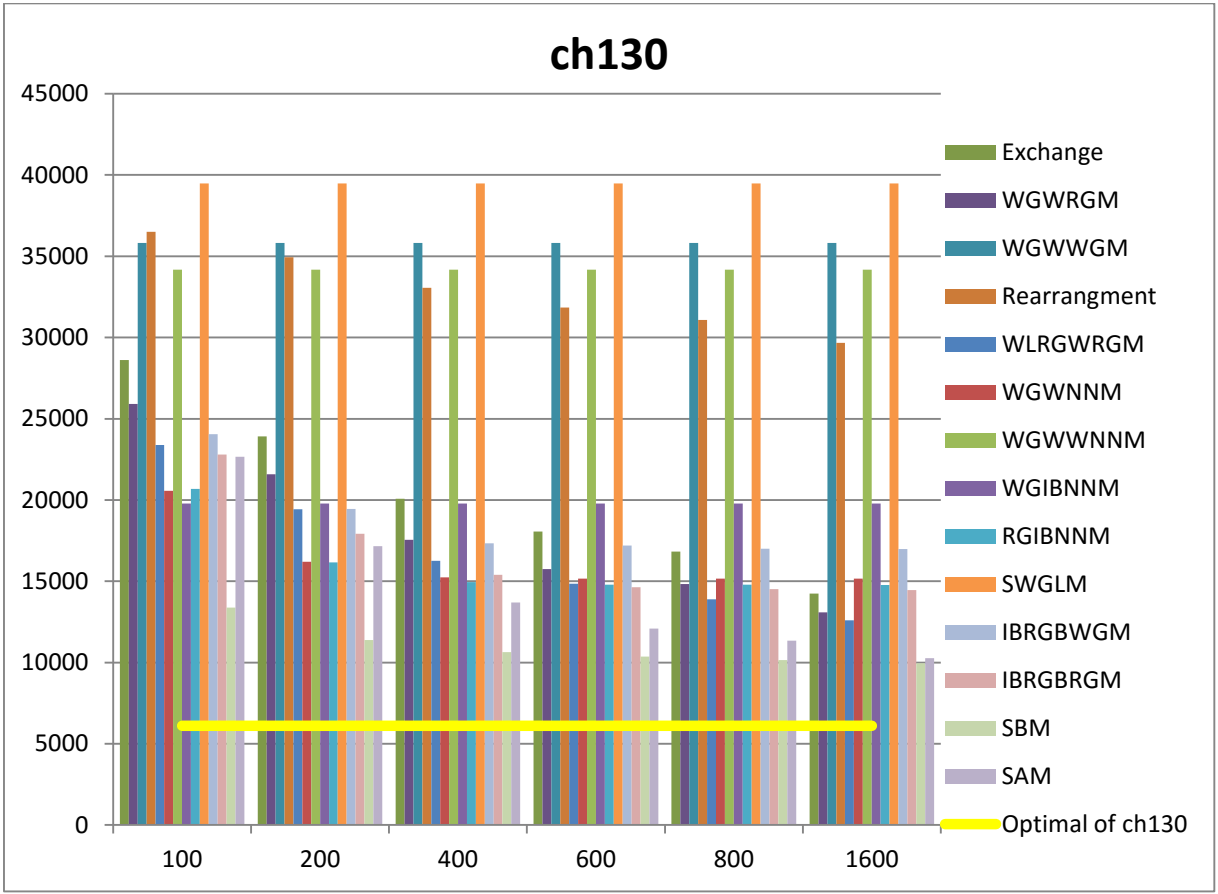

Fig. 9. Mutation's convergence to the minimum value, TSP (ch130)

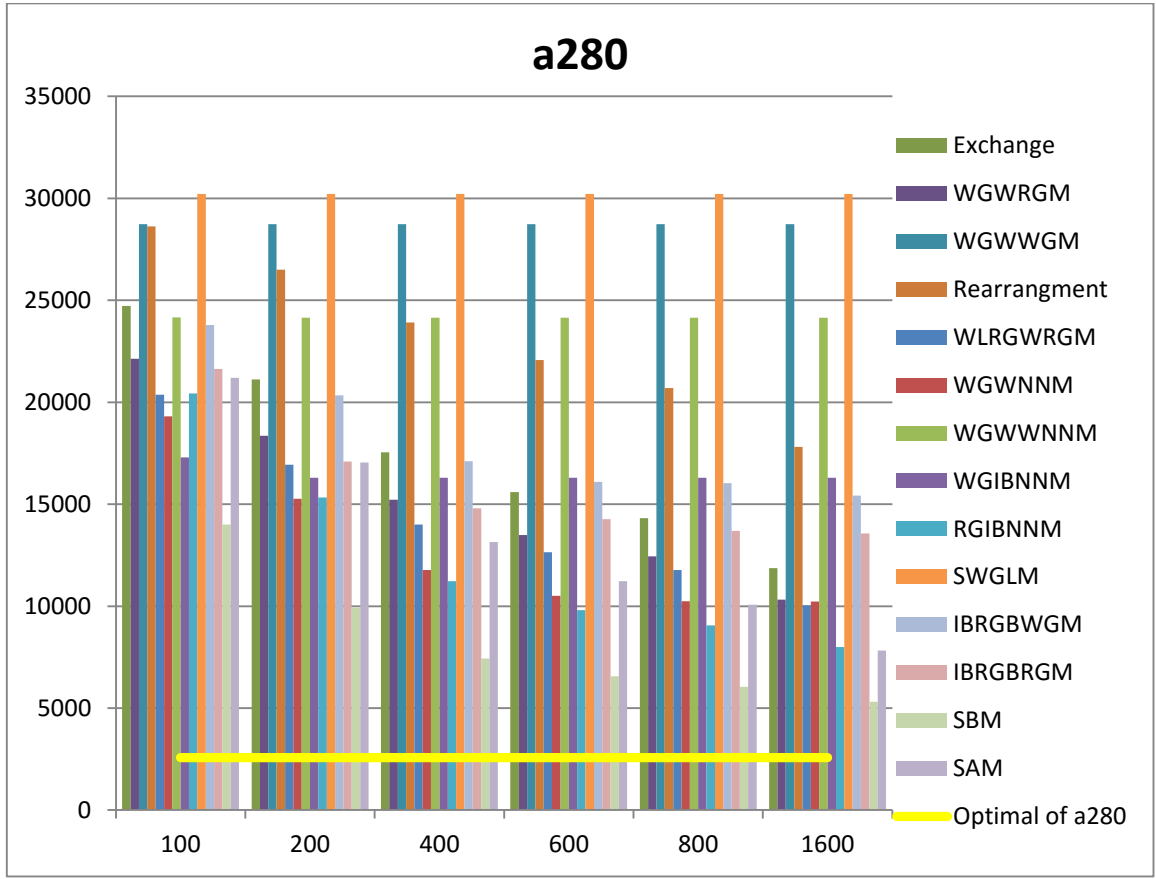

Fig. 10. Mutation's convergence to the minimum value, TSP(a280)

As can be seen from Figure 11, the best performance was recorded by the SBM algorithm. This showed faster convergence to the minimum value than any other mutation, followed by the SAM algorithm. At the level of mutations alone, RGIBNNM, followed by WLRGWRGM and WGWNNM in addition to WGWRGM mutations showed a better performance than the rest of the mutations. Because of the slow convergence of the SWGLM and WGWWGM mutations, they achieved the worst result.

The reason behind the good performance of the SBM is that it tries several mutations and chooses the best among them; however, this comes at the cost of time consumed. Although the SBM outperformed the SAM, SAM is still better than SBM in terms of time spent because SBM tries all mutations available and chooses the best, while SAM selects any one randomly. Moreover, the difference between the two results is sometimes not significant. The good performance of the SAM is due to using a different mutation each time, and this leads to an increase in the diversity of the solutions, and thus enhances the overall performance of the GA.

The second set of experiments attempted to measure the effectiveness of the SBM and SAM comparing to optimal solutions. These methods and all the proposed operators, in addition to the Exchange mutation and Rearrangement mutation, were tested using 13 TSP instances taken from the TSPLIB. They include a280, att48, berlin52, bier127, ch130, eil51, kroA100, pr76, pr144, u159, rat783, brd14051, and usa13509.

The genetic algorithm parameters that were selected were the same as in the first test; however, the recorded results were the average of the solutions at the last generation (1,600) after executing the algorithm 10 times (see Table 1).

TABLE I
RESULTS OF 13 TSP INSTANCES OBTAINED BY 14 MUTATION OPERATORS AFTER 1,600 GENERATIONS

| Mutation | a280 | att48 | berlin52 | bier127 | ch130 | eil51 | kroA100 | pr76 | pr144 | u159 | rat783 | brd14051 | usa13509 |
|---|---|---|---|---|---|---|---|---|---|---|---|---|---|
| **Exchange** | 11860 | 41749.4 | 9338.4 | 217739 | 13923 | 514.8 | 44815 | 169713 | 219250 | 133616 | 83155 | 36964078 | 1878070618 |
| **Rearrangement** | 17810 | 73119 | 15381 | 377025 | 29671 | 802.1 | 78546 | 272815 | 373603 | 208038 | 116095 | 35411256 | 1788855536 |
| **WGWRGM** | 10325 | 42221.8 | 10529 | 252213 | 13084 | 503.1 | 42259 | 168850 | 190946 | 122144 | 71748 | 41534181 | 2117784066 |
| **WGWWGM** | 28734 | 93108 | 20994 | 528898 | 35817 | 1050 | 119607 | 420047 | 660178 | 339365 | 165796 | 39752677 | 2035635792 |
| **WLRGWRGM** | 10043 | 43225.6 | 10714 | 262604 | 12606 | 524 | 44158 | 167912 | 200323 | 116924 | 68705 | 33441004 | 1681692076 |
| **WGWNNM** | 10233 | 46517.3 | 11075 | 338476 | 15172 | 589.9 | 50393 | 199048 | 234684 | 129658 | 58338 | 32788677 | 1613016352 |
| **WGWWNNM** | 24139 | 89746.5 | 19625 | 543930 | 34178 | 1073 | 107043 | 408988 | 557415 | 301068 | 143057 | 39139603 | 2065593522 |
| **WGIBNNM** | 16300 | 62576 | 14314 | 446290 | 19781 | 657.7 | 67283 | 234865 | 310768 | 199013 | 104155 | 30505628 | 1549822430 |
| **RGIBNNM** | 8000.2 | 49855 | 10193 | 225990 | 14777 | 551 | 47938 | 194527 | 213205 | 116383 | 56263 | 34597287 | 1735470678 |
| **SWGLM** | 30212 | 120925 | 23689 | 559770 | 39487 | 1275 | 139929 | 467464 | 696683 | 386194 | 166447 | 41361128 | 2126239629 |
| **IBRGBWGM** | 15416 | 66912.4 | 13009 | 328296 | 16987 | 659.7 | 66358 | 228258 | 321485 | 180738 | 101146 | 36218274 | 1853569535 |
| **IBRGBRGM** | 13562 | 45749.6 | 11378 | 256321 | 14465 | 583.4 | 48408 | 214855 | 261076 | 164734 | 68005 | 36058022 | 1822032402 |
| **SBM** | **5316.1** | **37575.8** | **8782.9** | **190978** | **9958.4** | **459.1** | 35063 | 147595 | **137256** | **78225** | **34777** | **27638514** | **1377597129** |
| **SAM** | 7830.7 | 38612.8 | 8875.3 | 201895 | 10262 | 469.9 | **33145** | **147369** | 142124 | 88452 | 59216 | 34314633 | 1708749204 |
| **Optimal** | **2579** | **10628** | **7542** | **118282** | **6110** | **426** | **21282** | **108159** | **58537** | **42080** | **8806** | **469385** | **19982859** |

TABLE II
RANKS OF MUTATION OPERATORS AFTER 1,600 GENERATIONS

| Mutation | a280 | att48 | berlin52 | bier127 | ch130 | eil51 | kroA100 | pr76 | pr144 | u159 | rat783 | brd14051 | usa13509 | Average |
|---|---|---|---|---|---|---|---|---|---|---|---|---|---|---|
| **Exchange** | 8 | 4 | 4 | 4 | 6 | 5 | 6 | 6 | 7 | 8 | 9 | 11 | 11 | 7 |
| **Rearrangement** | 12 | 12 | 12 | 11 | 12 | 12 | 12 | 12 | 12 | 12 | 12 | 13 | 12 | 12 |
| **WGWRGM** | 7 | 5 | 6 | 6 | 5 | 4 | 4 | 5 | 4 | 6 | 8 | 8 | 8 | 6 |
| **WGWWGM** | 14 | 14 | 14 | 13 | 14 | 13 | 14 | 14 | 14 | 14 | 14 | 15 | 14 | 14 |
| **WLRGWRGM** | 5 | 6 | 7 | 8 | 4 | 6 | 5 | 4 | 5 | 5 | 7 | 5 | 5 | 6 |
| **WGWNNM** | 6 | 8 | 8 | 10 | 9 | 9 | 9 | 8 | 8 | 7 | 4 | 4 | 4 | 7 |
| **WGWWNNM** | 13 | 13 | 13 | 14 | 13 | 14 | 13 | 13 | 13 | 13 | 13 | 12 | 13 | 13 |
| **WGIBNNM** | 11 | 10 | 11 | 12 | 11 | 10 | 11 | 11 | 10 | 11 | 11 | 3 | 3 | 10 |
| **RGIBNNM** | 4 | 9 | 5 | 5 | 8 | 7 | 7 | 7 | 6 | 4 | 3 | 7 | 7 | 6 |
| **SWGLM** | 15 | 15 | 15 | 15 | 15 | 15 | 15 | 15 | 15 | 15 | 15 | 14 | 15 | 15 |
| **IBRGBWGM** | 10 | 11 | 10 | 9 | 10 | 11 | 10 | 10 | 11 | 10 | 10 | 10 | 10 | 10 |
| **IBRGBRGM** | 9 | 7 | 9 | 7 | 7 | 8 | 8 | 9 | 9 | 9 | 6 | 9 | 9 | 8 |
| **SBM** | 2 | 2 | 2 | 2 | 2 | 2 | 3 | 3 | 2 | 2 | 2 | 2 | 2 | 2 |
| **SAM** | 3 | 3 | 3 | 3 | 3 | 3 | 2 | 2 | 3 | 3 | 5 | 6 | 6 | 3 |
| **Optimal** | 1 | 1 | 1 | 1 | 1 | 1 | 1 | 1 | 1 | 1 | 1 | 1 | 1 | 1 |

As can be seen in Table 1, results indicate the efficiency of the SBM algorithm in most of the problems, such as a280, rat87, berlin52, bier127, ch130, att48, pr144, u159, and eil51. It converges to the optimal faster than the exchange method, and the rest of the test data (instances) were outperformed by the SAM algorithm, such as pr76 and kroA100.

Considering methods that use one mutation only, the WGWRGM, WLRGWRGM and RGIBNNM performed better than other methods (see Table 2). The WGWRGM mutation was the best in three problems, eil51, kroA100 and pr144, and the RGIBNNM mutation was the best in three problems, a280, rat783 and u159. WLRGWRGM also showed convergence in the rest of the instances better than other methods. This method was the best in two problems. The Exchange mutation was the best in three problems, att48, berlin52 and bier127.

In these experiments, SWGLM showed weak performance, followed by WGWWGM which showed slow convergence to a minimum value. However, the importance of these operators has emerged in the diversity of the population, where both helped to achieve new areas for searching to be used by SAM and SBM.

The good performance of SBM was expected and not surprising, because SBM uses a number of mutations and chooses the best among them. Figure 12 shows the average selection probability for each mutation.

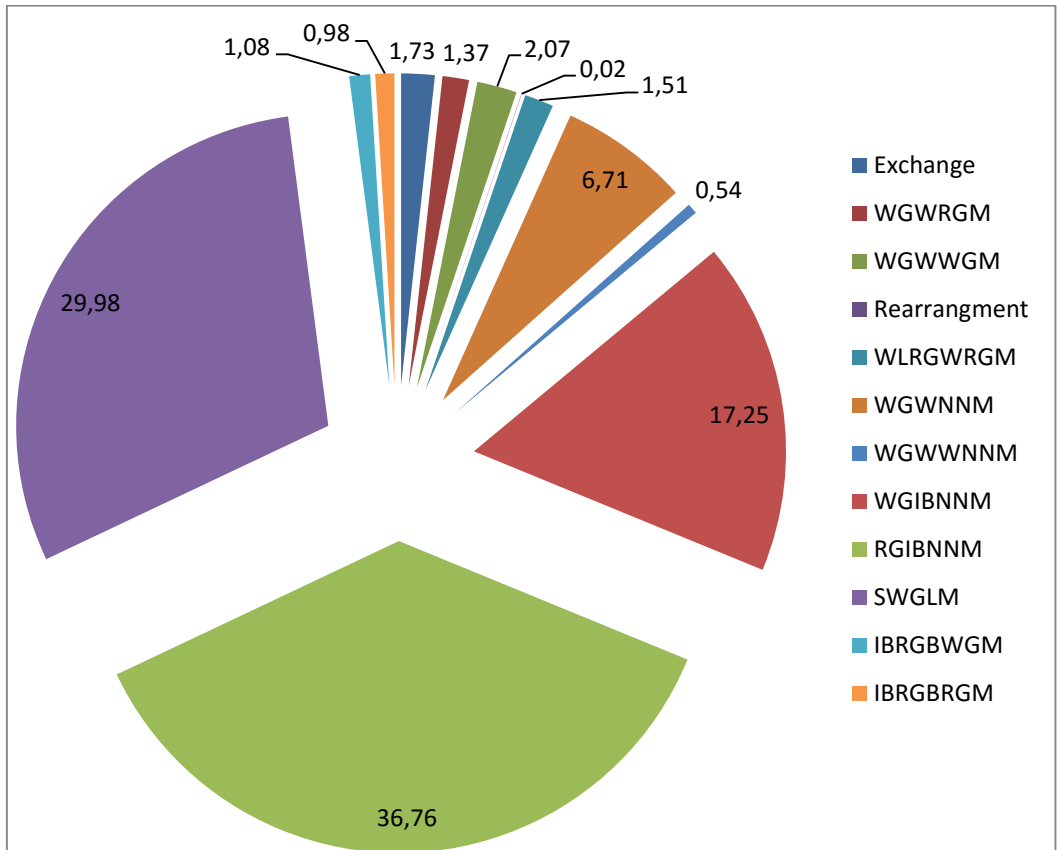

Fig. 11. The average selection probabilities for mutations used (all numbers are in percent).

As can be seen from Figure 12, the most selected mutation is the RGIBNNM, with an average probability of 36.76%. This is not surprising as this mutation performed, on average, better than most of the other methods (see Tables 1 and 2). Moreover, the least selected mutations were Rearrangement and WGWWNNM with 0.02% and 0.54% respectively. This was also not surprising as both mutations performed the weakest comparing to the other mutations. What is surprising is to have the SWGLM—the mutation of the weakest performance—selected by SBM with a probability of 29.98% ranked second. Perhaps the SWGLM contributes well to diversity, which increases the performance of the SBM.

It is interesting to note that the gap between SBM and SAM decreases as the number of generation increases (see Figures 9, 10 and 11), and sometimes the difference is not significant as in Figures 9 and 10 at generation 1,600. This shows that SAM is better than SMB in terms of time and accuracy if we used large number of generations. But if we want to use a small number of generations, SBM would be a better choice, as it converges to better solutions faster. Table 3 shows the average time consumed for each method for each TSP instance using single 3.06 GHz Pentium 4 CPU.

TABLE III
AVERAGE TIME (IN MILLISECONDS) CONSUMED BY EACH MUTATION AFTER 1,600 GENERATIONS

| Mutation | a280 | att48 | berlin52 | bier127 | ch130 | eil51 | kroA100 | pr76 | pr144 | u159 | rat783 | brd14051 | usa13509 |
|---|---|---|---|---|---|---|---|---|---|---|---|---|---|
| **Exchange** | **19843** | **14172** | 14008 | 15934 | 16562 | 13724 | 15825 | **14476** | **16749** | 17049 | 40598 | **514352** | 544840 |
| **WGWRGM** | 31481 | 21060 | 18984 | 21799 | 18424 | 15103 | 20179 | 20029 | 22077 | 25978 | 56772 | 666619 | 713398 |
| **WGWWGM** | 26840 | 18510 | 17732 | 19676 | 19662 | 14981 | 17647 | 14879 | 17573 | 21644 | 49813 | 630079 | 666275 |
| **Rearrangement** | 33122 | 15552 | 18872 | 20245 | 20756 | 18369 | 21096 | 17188 | 23967 | 22321 | 69531 | 1130893 | 1060503 |
| **WLRGWRGM** | 30126 | 17603 | 17969 | 22654 | 22736 | 17173 | 21103 | 19669 | 22784 | 22478 | 61324 | 1026570 | 741261 |
| **WGWNNM** | 37008 | 19360 | 16800 | 17980 | 30933 | 18177 | 21804 | 23597 | 28001 | 23240 | 78486 | 845265 | 809090 |
| **WGWWNNM** | 28452 | 17347 | 14415 | 17544 | 18470 | 13645 | 16967 | 15142 | 20581 | 20589 | 53550 | 792461 | 784913 |
| **WGIBNNM** | 28668 | 16937 | **13844** | 17994 | 20854 | 13900 | 18434 | 19831 | 21182 | 24417 | 65065 | 1005676 | 897441 |
| **RGIBNNM** | 24139 | 16860 | 17442 | 21564 | 20274 | 14448 | 19486 | 16354 | 20062 | 22487 | 44072 | 642389 | 498415 |
| **SWGLM** | 28409 | 17343 | 14772 | 17208 | 21388 | 16581 | 17434 | 18373 | 22412 | 23081 | 57737 | 800739 | 774224 |
| **IBRGBWGM** | 27982 | 17350 | 13944 | 16224 | 18153 | 13586 | 17234 | 15132 | 18850 | 24867 | 52849 | 668979 | 601137 |
| **IBRGBRGM** | 27633 | 19907 | 15241 | **15457** | 16195 | **13266** | **14912** | 17308 | 18921 | **15980** | **38180** | 613229 | **465612** |
| **SBM** | 100975 | 30443 | 29452 | 53541 | 58931 | 30706 | 48506 | 40462 | 65555 | 66766 | 260287 | 8430944 | 7957288 |
| **SAM** | 27337 | 16214 | 15003 | 16746 | 18462 | 16975 | 19389 | 16155 | 20636 | 21133 | 58314 | 1048040 | 1445483 |
| **SBM/SAM** | 4 | 2 | 2 | 3 | 3 | 2 | 3 | 3 | 3 | 3 | 4 | 8 | 6 |

As can be seen from Table 3, the consumed time by the individual mutations, the first 12 mutations, is not significantly different. However, we find that the IBRGBRGM and the Exchange mutations consumed slightly less time than the others. This is because they do not need to do any special treatments to the mutated chromosome, such as finding the worst gene, which justifies the increase in the time consumed by the other mutations.

In addition, it is expected that SBM would consume more time than the others. Compared to the SAM, the SBM consumes at least double the time consumed by the SAM (see the last row in Table 3). This triggers the question: do the results of the SMB justify its high consumption of time? The answer depends mainly on the TSP instance itself, as can be seen from Table 3. The larger the number of instances the higher the time consumed by the SBM, and vice versa. Having the SBM converge faster than any other method for a better solution, with little time consumed when applied in small TSP instances makes the SBM a better choice. In the case of large TSP instances (greater than 200 cities), we do not recommend the use of SBM but instead recommend the use of SAM, as the results of SBM are not significantly better than the SAM, and the consumed time is much higher.

Although the aim of this paper is not to find the optimal solution for TSP instances, the solutions of the proposed algorithms were close to optimal solutions in some cases (e.g. columns 4 and 7 of Table 1), and none could achieve an optimal solution. Perhaps employing crossover operators and increasing the number of generations would enhance the solutions of the proposed methods. This shows the importance of using appropriate parameters along with mutation (such as population size, crossover ratio, number of generations, etc.), due to the effective impact of their convergence to an optimal or near optimal solution. Therefore, it is unfair to compare our findings to state-of-the-art GAs, as we just investigated the efficiency of the proposed mutations, and whether of to use a single mutation or more at each generation.

## V. CONCLUSIONS

We have proposed several mutation methods—WGWRGM, WGWWGM, WLRGWRGM, WGWNNM, WGWWNNM, WGIBNNM, RGIBNNM, SWGLM, IBRGBWGM and IBRGBRGM—to enhance the performance of GA while searching for near optimal solutions for the TSP, in addition to proposing two selection approaches—SBM and SAM. Several experiments were conducted to evaluate those methods on several TSP problems, which showed the efficiency of some of the proposed methods over the well-known Exchange mutation and Rearrangement mutations. Some of the proposed mutations can be used for other problems with some modifications and not only oriented to the TSP problem, such as the knapsack problem. Here the concept of the worst gene is defined by its value-over-weight ratio, except for those which uses the distance and the nearest neighbor approaches.

The results of the experiments conducted for this study also suggest that using more than one mutation method in the GA is preferable, because it allows the GA to avoid local optima; the proposed SBM and SAM strategies enhance the performance of the GA. This approach, using more than one mutation for GA, is supported [30], [31] and [34].

For the use of each mutation alone, some mutations showed better performance than others, and this does not mean that the rest of the mutations had been proven to fail. Even those with the weakest performance can be effective in dealing with other problems because every problem has a different search space. In this work, we found them effective in SBM and SAM, where they encouraged diversity and hence increased the efficiency of both algorithms.

Our future work will include the development of some types of new crossovers, using the same approaches, i.e. trying more than one crossover each time to support the proposed approaches [41], and attempting to further enhance the performance of GA. Additionally we will apply the proposed methods to different problems using different benchmark data.

ACKNOWLEDGMENT

The first author made part of his contribution to the work presented in this paper while he was a Visiting Research Fellow at the Sarajevo School of Science and Technology (www.ssst.edu.ba). He thanks the University for hosting his visit and for all the support and assistance afforded to him during the visit.

17

[41] A. B. A. Hassanat and E. Alkafaween, "On Enhancing Genetic Algorithms Using New Crossovers," *International Journal. of Computer Applications in Technology (IJCAT),* 2016.

[42] Reinelt and Gerhard, "TSPLIB. University of Heidelberg," 1996. [Online]. Available: http://comopt.ifi.uni-heidelberg.de/software/TSPLIB95/. [Accessed 17 9 2015].